\documentclass[letterpaper]{article} 
\usepackage{aaai2026}  
\usepackage{times}  
\usepackage{helvet}  
\usepackage{courier}  
\usepackage[hyphens]{url}  
\usepackage{graphicx} 
\usepackage{natbib}  
\usepackage{caption} 
\usepackage[utf8]{inputenc} 
\usepackage{booktabs}       
\usepackage{amsfonts}       
\usepackage{nicefrac}       
\usepackage{microtype}      
\usepackage[dvipsnames]{xcolor}         
\usepackage{svg}
\usepackage{amsmath}
\usepackage{makecell}
\usepackage{pifont}
\definecolor{mygreen}{HTML}{97d077}
\definecolor{lightred}{HTML}{DB8686}
\definecolor{lightred2}{HTML}{c99f9f}

\title{Unlocking Efficient Vehicle Dynamics Modeling\\via Analytic World Models}
\author{
    Asen Nachkov\textsuperscript{\rm 1}, Danda Pani Paudel\textsuperscript{\rm 1}, Jan-Nico Zaech\textsuperscript{\rm 1}, Davide Scaramuzza\textsuperscript{\rm 2}, Luc Van Gool\textsuperscript{\rm 1} 
}
\affiliations{
    \textsuperscript{\rm 1}INSAIT, Sofia University “St. Kliment Ohridski”, Sofia, Bulgaria \\
    \textsuperscript{\rm 2}University of Zurich, Zurich, Switzerland \\

}

\usepackage{algorithm}
\usepackage{algorithmic}
\usepackage{amsmath} 
\usepackage{graphicx}
\usepackage{multirow}
\usepackage{tabularx}

\usepackage{booktabs} 
\usepackage{array} 
\usepackage{caption} 
\usepackage{pifont}
\usepackage{makecell} 
\usepackage{colortbl} 
\usepackage{siunitx} 

\newcommand{\grayline}{\arrayrulecolor{gray}\hline\arrayrulecolor{black}}

\usepackage[title]{appendix}

\begin{document}

\maketitle

\begin{abstract}
Differentiable simulators represent an environment's dynamics as a differentiable function. Within robotics and autonomous driving, this property is used in Analytic Policy Gradients (APG), which relies on backpropagating through the dynamics to train accurate policies for diverse tasks. Here we show that differentiable simulation also has an important role in world modeling, where it can impart predictive, prescriptive, and counterfactual capabilities to an agent. Specifically, we design three novel task setups in which the differentiable dynamics are combined within an end-to-end computation graph not with a policy, but a state predictor. This allows us to learn relative odometry, optimal planners, and optimal inverse states. We collectively call these predictors Analytic World Models (AWMs) and demonstrate how differentiable simulation enables their efficient, end-to-end learning. In autonomous driving scenarios, they have broad applicability and can augment an agent's decision-making beyond reactive control.
\end{abstract}

\section{Introduction}
\label{sec: intro}

Differentiable simulation (DiffSim) has emerged as a powerful tool to train controllers and predictors across different domains like physics \cite{holl2020learning}, graphics \cite{laine2020modular}, and robotics \cite{hu2019difftaichi, degrave2019differentiable}. At its core, it is the ability to differentiate through an environment's dynamics, which in turn  allows us to embed the environment within a broader computational graph. Training in such an end-to-end loop involves differentiating both through the forward passes of any modules involved, as well as the dynamics themselves. 

An immediate application for this is policy learning, with the corresponding method called Analytic Policy Gradients (APG) \cite{nachkov2024autonomous}. For autonomous vehicle (AV) simulation, the dynamics represent the equations of motion that evolve a vehicle's state from one timestep to the next, under a specific action. To learn an optimal policy, APG  repeatedly rolls out a trajectory from the current policy, and supervises it with a reference expert one. During backpropagation gradients from the difference between each realized and expert state pass through the dynamics, reach the policy and update its weights.

\begin{figure}[t]
    \centering
    \includegraphics[width=1\columnwidth]{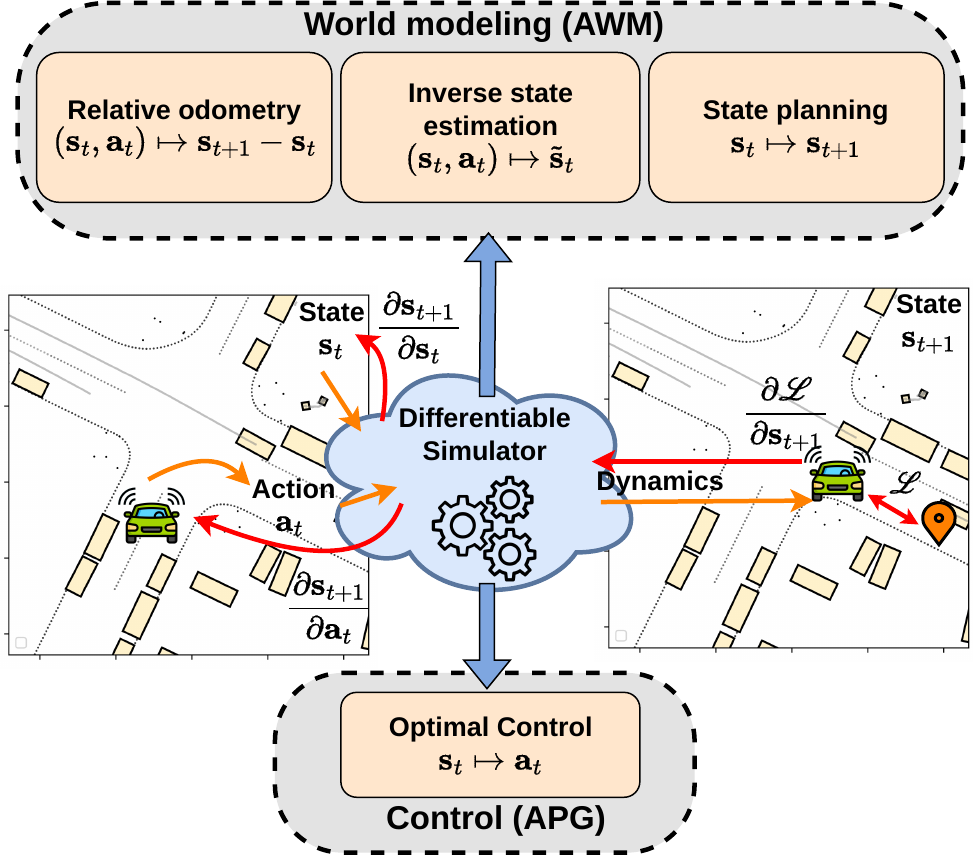}
    \captionsetup{belowskip=-0.55cm, aboveskip=0.2cm}
    \caption{\textbf{Differentiable simulation for world modeling.}
    Previously, differentiable simulation has been used to train controllers using analytic policy gradients (bottom). Our contribution is in applying it for learning relative odometry, state planning, and inverse state estimation (top).}
    \label{fig: teaser}
\end{figure}

Yet, an important question is whether the applications of differentiable simulation end with policy learning. A fundamental task for any agent, especially a driving vehicle, is world modeling -- predicting different states of interest like next states, desired states or counterfactual states. For a self-driving vehicle such world modeling is crucial to assess the effects of its actions as it navigates around other traffic participants in densely populated scenarios. And similar to control, predicting the world requires understanding its dynamics, which is where world modeling connects with differentiable simulation. Thus, in this work we are interested in designing an APG-style training setup not for policy learning, but for world modeling. Our main claim is that differentiable simulation allows us to efficiently obtain accurate predictors for diverse world modeling tasks, whereas without it, one would need to rely on costly trial-and-error search. 

To understand the efficiency that DiffSim provides, we consider first that in APG the gradients of the dynamics automatically guide the policy toward the optimal actions that reproduce the expert states. There is no search involved, it is all end-to-end, unlike when the environment is treated as a black box, in which case the optimal actions have to be learned through trial-and-error \cite{sutton1999policy}. Second, APG minimizes a training loss not in the action space, as methods like behavior cloning do, but in the state space. This allows any nonlinear effects in the dynamics to condition the policy, causing it to learn more physically-consistent features. We aim to capture these benefits in our world models.

The concept of a world model is nuanced, as there are different ways to understand the effect of one's own actions. Fig. \ref{fig: teaser} shows our approach, which uses the differentiability of the simulator to formulate three task setups related to world modeling. First, the effect of an agent's action could be understood as the difference between the agent's next state and its current state. If a vehicle's state consists of its position, yaw, and velocity, then this setup has an odometric interpretation, asking  the question \emph{``Where will the agent go?"}. Second, an agent could predict a desired next state to visit, which is a form of state planning. It asks the question \emph{``Where should it go?"}. Third, we can ask \emph{``Given an action in a particular state, what should that state be so this action is optimal?"}, which is another form of world modeling but also an inverse problem, effectively asking the counterfactual \emph{``Where should the agent have been?"}. All these tasks can be solved using trial-and-error approaches, as in RL, where the agent has to explore and search. Yet, our DiffSim designs solve them efficiently, without search, from direct supervision. Similar to APG, we call the corresponding predictors Analytic World Models (AWMs) to indicate they are trained using DiffSim and a state is predicted, instead of an action.

We note that some world modeling tasks like simple next-state prediction do not need a differentiable simulator to be solved efficiently. These are tasks in which the prediction targets come from the same policy as the one that was used to collect the training transitions. For any other ``off-policy" tasks, DiffSim is invaluable, as we show in Sec. \ref{sec: method}.

In our experiments we use the Waymax \cite{gulino2024waymax} autonomous driving simulator, which is fully differentiable, vectorizable, functional in design, and GPU-accelerated. It is also data-driven, as it replays scenarios from the realistic, large-scale Waymo Open Motion Dataset (WOMD) \cite{ettinger2021large}. Specifically, our AWMs are trained for the freely controlled ego-vehicle, whose realized trajectories can deviate from the historical ones, while all other agents evolve according to their historical motion in the dataset. The trained agent can reason about the world dynamics with its AWMs, is lightweight, with just over 6M parameters, and runs in real-time. Training the AWMs is similarly cheap.

Our work covers diverse experimental settings. We aim to highlight the breadth of the application of DiffSim and to that end we purposefully provide separate, independent experiments for the AWM tasks. At test time the agent only has access to its learned modules. Our contributions are:
\begin{itemize}
    \item We propose Analytic World Models (AWMs) -- different state predictors  trained with differentiable simulation.
    \item We evaluate them in diverse settings, improving over relevant baselines and previous works.
\end{itemize}
\section{Related Work}
\label{sec: related_work}

Our work uses differentiable simulation for world modeling, within an AV setting. We cover the relevant context below.

\textbf{Differentiable simulation.} Differentiable simulators have grown in popularity because they allow one to solve ill-posed inverse problems related to the dynamics. As examples, an object's physical parameters like mass, friction, and elasticity could be estimated directly from videos and real-world experiments \cite{de2018end, murthy2020gradsim, geilinger2020add}, or simulations of soft material cutting could enable precise calibration and policy learning \cite{heiden2021disect}. Simulations can be parallelized across accelerators to enable efficient scaling of problem and experiment sizes \cite{xu2022accelerated, warp2022, freeman2021brax}. Within the field of robotics, differentiable simulation is used extensively, especially for training robotic policies in physically-realistic settings \cite{newbury2024review, lutter2021differentiable, toussaint2018differentiable}. The focus has often been on object manipulation \cite{li2023dexdeform, xu2021end, lin2022diffskill} which requires having differentiable contact models for detecting collisions. Analytic policy gradients (APG) has been used to train policies for trajectory tracking and navigation in quadrotors and fixed-wing drones \cite{wiedemann2023training}, quadruped locomotion \cite{song2024learning}, and for quadrotor control from visual features \cite{heeg2024learning}, yielding strong results. 

\textbf{Autonomous vehicles}. For AVs simulators are crucial. Photorealistic simulators \cite{dosovitskiy2017carla, martinez2017beyond, li2023scenarionet} have been used for visual control \cite{codevilla2018end, zhang2019vr}, with less focus on differentiable motion ones \cite{lavington2024torchdriveenv}. Some are differentiable but lack expert actions \cite{lavington2024torchdriveenv}, others are lacking acceleration support, crucial for large scale training \cite{sun2022intersim, li2022metadrive, vinitsky2022nocturne}. Waymax \cite{gulino2024waymax} is a differentiable data-driven simulator for vehicle motion. It allows one to simulate trajectories instantiated from real driving scenarios and compare them to the historical human drivers' motion, which are considered ground-truth. It also provides inverse kinematics -- computing the action that transfers the simulator to a particular next state, which is valuable for our AWMs.

\textbf{Baseline}. The work most relevant to ours is Analytic Policy Gradients (APG) applied for vehicle motion \cite{nachkov2024autonomous}. It trains near-optimal policies in a supervised manner, relying on the differentiability of the Waymax simulator. Using a recurrent architecture, the model selects actions autoregressively from the observed agent locations, nearest roadgraph points, traffic lights, and goal heading. At training time the model learns to select those actions that would bring the simulated trajectory as close as possible to the expert trajectory. Due to the RNN architecture, the derivatives of the dynamics from each timestep mix with those of the RNN hidden state and propagate backwards until the start of the trajectory. Our goal is to generalize this approach to world modeling.

\textbf{Model-based methods.} Action-selection using world models \cite{schrittwieser2020mastering, ha2018world, moerland2023model} is a common problem with two main approaches: model-predictive control (MPC) \cite{bertsekas2012dynamic} and Dyna-style imagination \cite{sutton1991dyna}. With MPC \cite{arroyo2022reinforced, romero2024actor}, one starts with a random policy from which actions are sampled and evaluated. Then, the policy is repeatedly refit on only the best trajectories, from which new trajectories are sampled. Eventually, an aggregated action from the best trajectories is selected and executed. Being closed-loop, this strategy is repeated at every timestep. To assess the possibility of using our world models for action selection beyond reactive settings, in Sec. \ref{subsec: mpc} we perform an experiment where we adopt MPC as the main action selection framework, while the learned AWMs are used to predict and score the trajectories.

\section{Method}
\label{sec: method}

\textbf{Notation.} We represent the current simulator state with $\mathbf{s}_t$, the current action with $\mathbf{a}_t$, the log (= human expert = ground-truth = reference) state with $\hat{\mathbf{s}}_t$, the log action with $\hat{\mathbf{a}}_t$. The simulator is a function $\text{Sim}: \mathcal{S} \times \mathcal{A} \rightarrow \mathcal{S}$, with $\text{Sim}(\mathbf{s}_t, \mathbf{a}_t) = \mathbf{s}_{t + 1}$, where the set of all states is $\mathcal{S}$ and that of the actions $\mathcal{A}$. The mapping $\text{InvKin}(\mathbf{s}_t, \mathbf{s}_{t+1}) = \mathbf{a}_t$ is called inverse kinematics and produces the action that transfers one state into another. We denote policies as $\pi_\theta$ and world models as $f_\phi$, with $\theta$ and $\phi$ representing their parameters.

\textbf{Strategy}. To understand how DiffSim unlocks the efficient solving of diverse world modeling tasks we will show it is difficult to obtain similar predictors without DiffSim. Training there needs to happen by collecting transitions to train on, and obtaining the labels to supervise with. The main questions are: (i) how will the action selection during training happen, i.e. how to perform the rollout, and (ii) what variables do we supervise with. We will show that without DiffSim, key variables cannot be obtained, and hence one would need to rely on trial-and-error training, which is sample-inefficient.

\subsection{Preliminaries -- APG}
\label{subsection: apg}

It is shown in \cite{nachkov2024autonomous} that a differentiable simulator can turn the unsupervised search problem of optimal policy learning into a supervised one. Here the policy $\pi_\theta$ produces an action $\mathbf{a}_t$ from the current state, which is executed in the environment to obtain the next state $\mathbf{s}_{t+1}$. Comparing it to the reference trajectory $\hat{\mathbf{s}}_{t+1}$ produces a loss, whose gradient is backpropagated through the simulator and back to the policy:
\vspace{-0.1cm}
\begin{equation} \label{eq:apg}
\begin{aligned}
\min_\theta {\Big\lVert \text{Sim} \big(\mathbf{s}_t, \pi_\theta (\mathbf{s}_t) \big) - \hat{\mathbf{s}}_{t+1} \Big\rVert}_2^2.
\end{aligned}
\end{equation}
The key gradient here is that of the next state with respect to the current agent actions $\frac{\partial \mathbf{s}_{t+1}}{\partial \mathbf{a}_t}$. The loss is minimized whenever the policy outputs an action equal to the inverse kinematics $\text{InvKin}(\mathbf{s}_t , \hat{\mathbf{s}}_{t+1})$, which is also what the policy implicitly learns. To obtain similar supervision without DiffSim, one would need to supervise the policy with the inverse kinematic actions, which are unavailable if the environment is considered a black box. Hence, this is an inverse problem that is not efficiently solvable without access to a known environment, in this case to provide inverse kinematics.

\begin{figure}[t]
    \centering
    \includegraphics[width=1\columnwidth]{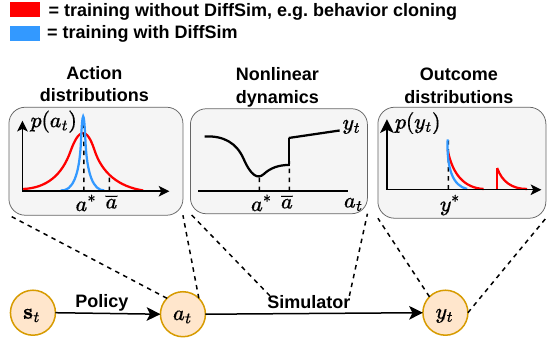}
    \captionsetup{belowskip=-0.4cm, aboveskip=0cm}
    \caption{\textbf{The benefits of differentiable simulation.} Methods that do not use DiffSim, e.g. behavior cloning, shown in \textcolor{red}{red}, are trained to minimize a loss in the action space. If the dynamics are nonlinear (here with a jump at the action $\bar{a}$), the distribution of the outcome could be bad. DiffSim-based methods (\textcolor{NavyBlue}{blue}) minimize a loss directly in the outcome space and the learned action distributions are tighter.}
    \label{fig: diffenv_diagrams}
\end{figure}

Furthermore, DiffSim is beneficial in APG because the loss in Eqn. \ref{eq:apg} is applied in the state space, as opposed to the action space. Fig. \ref{fig: diffenv_diagrams} shows schematically the effects of this. 

\subsection{Relative Odometry}
\label{subsection: dynamics}

In this setting a world model $f_\phi^O : \mathcal{S} \times \mathcal{A} \rightarrow \mathcal{S}$ predicts the next state $\mathbf{s}_{t+1}$ from the current state-action pair $(\mathbf{s}_t, \mathbf{a}_t)$. A differentiable simulator is not strictly needed to learn such a predictor. One can obtain $(\mathbf{s}_t, \mathbf{a}_t, \mathbf{s}_{t+1})$ tuples by rolling out a policy and then supervising the predictions with the next state $\mathbf{s}_{t+1}$. However, this would not utilize the available dynamics in any way. Hence we provide a formulation for bringing the simulator into an end-to-end training loop:
\begin{equation} \label{eq:world_model_with_inverse}
\begin{aligned}
\min_\phi {\Big\lVert \text{Sim}^{-1} \big(f^O_\phi(\mathbf{s}_t, \mathbf{a}_t), \mathbf{a}_t \big) - \mathbf{s}_{t} \Big\rVert}_2^2.
\end{aligned}
\end{equation}
Here, the world model $f_\phi^O$ takes $(\mathbf{s}_{t}, \mathbf{a}_t)$ and returns a next-state estimate $\tilde{\mathbf{s}}_{t+1}$. We then feed it into an inverse simulator $\text{Sim}^{-1}$ which is a function with the property that $\text{Sim}^{-1}( \text{Sim}(\mathbf{s}_t, \mathbf{a}_t), \mathbf{a}_t) = \mathbf{s}_t$. This output is compared with the current $\mathbf{s}_t$. The loss is minimized when $f_\phi^O$ predicts exactly $\mathbf{s}_{t+1}$, thus becoming a predictor of the next state, conditional on the provided action $\mathbf{a}_t$.

We implement the inverse simulator for the bicycle dynamics in Waymax. They embody the most realistic nonlinear vehicle motion available in the simulator. The velocities $v_x$ and $v_y$ are tied to the yaw angle $\alpha$ of the agent through the relationship $v_x = v \cos \alpha$ and $v_y = v \sin \alpha$, where $v$ is the current speed. However, at the first simulation step, due to WOMD being collected with noisy estimates of the agent state parameters, the relationships between $v_x$, $v_y$, and $\alpha$ do not hold. Thus, the inverse simulator produces incorrect results for the first timestep.

For this reason, we change the design used in the experiments to one that only requires access to a forward simulator:
\begin{equation} \label{eq:world_model_without_inverse}
\begin{aligned}
\min_\phi {\Big\lVert \text{Sim} \big(\mathbf{s}_{t+1} - f_\phi^O(\mathbf{s}_t, \mathbf{a}_t), \mathbf{a}_t \big) - \mathbf{s}_{t + 1} \Big\rVert}_2^2,
\end{aligned}
\end{equation}
where $f_\phi^O$ predicts the relative state difference that executing $\mathbf{a}_t$ will bring to the agent. One can verify that the loss is minimized if the prediction is equal to $\mathbf{s}_{t+1} - \mathbf{s}_t$. This can still be interpreted as a world model where $f_\phi^O$ learns to estimate how an action would change its relative state. Since the time-varying elements of the agent state consist of $(x, y, v_x, v_y, \alpha)$, this world model has a clear relative odometric interpretation. Learning such a predictor without a differentiable simulator will prevent the gradients of the environment dynamics from mixing with those of the network, which is suboptimal. 

\textbf{Inverse dynamics and inverse kinematics}. Given tuples $(\mathbf{s}_t, \mathbf{a}_t, \mathbf{s}_{t + 1})$, one can learn inverse dynamics $(\mathbf{s}_{t+1}, \mathbf{a}_t) \mapsto \mathbf{s}_t$ and inverse kinematics $(\mathbf{s}_{t}, \mathbf{s}_{t+1}) \mapsto \mathbf{a}_t$ without a differentiable simulator \cite{pathak2017curiosity}, but this is still completely agnostic in terms of how the data was generated. Formulations that involve the simulator are also possible. We do not list them here because they are similar to Eqn.~\ref{eq:world_model_without_inverse}.

\subsection{Optimal Planners}
\label{subsection: optimal_planners}

We call the mapping $f_\phi^P: \mathcal{S} \rightarrow \mathcal{S}$ with $\mathbf{s}_t \mapsto \mathbf{s}_{t+1} - \mathbf{s}_{t}$ a planner because it plans out the next state to visit from the current one. Unlike a policy, which selects an action without explicitly knowing the next state, the planner does not execute any actions. Until an action is executed, its output is inconsequential. We consider the problem of learning an optimal planner with respect to the expert trajectories. With a differentiable simulator we can formulate the problem as:
\begin{equation} \label{eq:optimal_planner}
\begin{aligned}
\min_\phi {\Big\lVert \text{Sim} \Big(\mathbf{s}_t, \text{InvKin}\big(\mathbf{s}_t, \mathbf{s}_t + f_\phi^P(\mathbf{s}_t) \big) \Big) - \hat{\mathbf{s}}_{t + 1} \Big\rVert}_2^2.
\end{aligned}
\end{equation}
Here, $f_\phi^P$ predicts the next state to visit as an offset to the current one. The action that reaches it is obtained using the inverse kinematics. After executing that action we directly supervise with the optimal next state. The gradient of the loss goes through the simulator, the inverse kinematics, and finally through the state planner network. Note that with a black box environment we can still supervise the planner directly with $\hat{\mathbf{s}}_{t+1}$, but a black box does not provide any inverse kinematics, hence there is no way to perform trajectory rollouts, unless with a separate behavioral policy.

\subsection{Inverse Optimal State Estimation}
\label{subsection: inverse_state_estimation}

We now consider the following task \emph{``Given $(\mathbf{s}_t, \mathbf{a}_t)$, find an alternative state $\tilde{\mathbf{s}}_t$ for the current timestep $t$ where taking action $\mathbf{a}_t$ will lead to an optimal next state $\hat{\mathbf{s}}_{t+1}$"}. This represents the counterfactual statement ``Had the agent been in $\tilde{\mathbf{s}}_t$, then the action $\mathbf{a}_t$ would have been optimal". We formulate the learning objective as
\begin{equation} \label{eq:inverse_state_estimation}
\begin{aligned}
\min_\phi {\Big\lVert \text{Sim} \big(\mathbf{s}_t + f_\phi^I(\mathbf{s}_t, \mathbf{a}_t), \mathbf{a}_t \big) - \hat{\mathbf{s}}_{t + 1} \Big\rVert}_2^2,
\end{aligned}
\end{equation}
where $f_\phi^I$ needs to estimate the effect of the action $\mathbf{a}_t$ and predict a new state $\tilde{\mathbf{s}}_t$, relatively to the current state $\mathbf{s}_t$, such that after executing $\mathbf{a}_t$ in it, the agent reaches $\hat{\mathbf{s}}_{t + 1}$. The loss is minimized if $f_\phi^I$ predicts $\tilde{\mathbf{s}}_t - \mathbf{s}_t$. The key gradient, as in Eqns.~\ref{eq:world_model_with_inverse}, \ref{eq:world_model_without_inverse}, and \ref{eq:optimal_planner}, is that of the next state with respect to the current state. Given the design of the Waymax simulator, these gradients are readily-available.

Consider solving this task with a black box environment. To do so, one would need to supervise the prediction  $f_\phi^I(\mathbf{s}_t, \mathbf{a}_t)$ with some particular state $\tilde{\mathbf{s}}_t - \mathbf{s}_t$, with $\tilde{\mathbf{s}}_t$ being unknown. By definition $\tilde{\mathbf{s}}_t = \text{Sim}^{-1}(\hat{\mathbf{s}}_{t+1}, \mathbf{a}_t)$, which is unobtainable since under a black box environment assumption, $\text{Sim}^{-1}$ is unavailable. Hence, this is another inverse problem which is not efficiently solvable unless we are given more information about the environment's inverse function.

The utility of this task is in providing a \emph{confidence} measure to an action. If the prediction of $f_\phi^I$ is close to $\mathbf{0}$, then the agent is relatively certain that the action $\mathbf{a}_t$ is close to optimal. Likewise, a large prediction from $f_\phi^I$ indicates a belief that the state $\mathbf{s}_{t+1}$ after $(\mathbf{s}_t, \mathbf{a}_t)$ will be far from the expert one. 

\textbf{Summary of DiffSim strengths}. Table \ref{table: diffsim_advantages} summarizes how training such predictors would work without DiffSim. It highlights that with a black box environment it is not possible to explicitly obtain the target variables with which to supervise the predictors. Therefore, one has to rely on exploration and trial-and-error search. Conversely, the white-box nature of a differentiable simulator allows us to (i) sidestep the trial-and-error learning, unlocking efficient training, and (ii) incorporate the dynamics within an end-to-end training loop, which helps predictors learn more physically-consistent features. 
\begin{table}[t]
    \small
  \centering
  \begin{tabular}{@{} l l l l @{}}
    \toprule[1.5pt]
    \textbf{Type} & \textbf{Predict} & \makecell[tl]{\textbf{Action}\\\textbf{in Rollout}} & \makecell[tl]{\textbf{Supervise}\\\textbf{with}} \\
    \midrule[1pt]
    APG & Opt. action $\mathbf{a}_t$ & From policy & \cellcolor{lightred} inv. kin. \\
    \midrule
    \multirow{3}{*}{AWM}
        & Next state $\mathbf{s}_{t+1}$ & From policy & $\mathbf{s}_{t+1}$             \\
        & Opt. next state & \cellcolor{lightred}From inv. kin. & $\hat{\mathbf{s}}_{t+1}$        \\
        & Inv.\ opt.\ state  & From policy  & \cellcolor{lightred}$\text{Sim}^{-1}$  \\
    \bottomrule[1.5pt]
  \end{tabular}
  \small
  \captionsetup{belowskip=-0.5cm, aboveskip=0.15cm}
  \caption{\textbf{Characteristics of learning inverse dynamics predictors without DiffSim}. A black box environment is missing the computations highlighted in \textcolor{lightred}{red} and training without DiffSim would require sample-inefficient search.}
  \label{table: diffsim_advantages}
\end{table}
\section{Experiments}
\label{sec: experiments}

To cover the range of applications of DiffSim, we provide a comprehensive suite of experiments. We evaluate the four differentiable simulation tasks separately in Sec. \ref{subsec: awms_eval}. In Sec. \ref{subsec: mpc} we show joint evaluation based on model-predictive control.

\textbf{AWM agent architecture}. The inputs to the driving agent include the locations of all other traffic participants, the nearest roadgraph points, the traffic lights, the ego vehicle's own velocity, and any route features (heading angle towards destination, or final $(x, y)$ waypoint) to mark the goal destination. A scene encoder extracts features from all these modalities and fuses them. A recurrent component evolves these state features in time. The three AWMs and the policy are implemented as four parallel heads on top of these features. They do not share parameters (here indicated generically as $\theta$ and $\phi$).  The policy is trained using Analytic Policy Gradients (APG) and its collected data is used to train the AWMs. The loss functions are Eqns. \ref{eq:apg}, \ref{eq:world_model_without_inverse}, \ref{eq:optimal_planner}, and \ref{eq:inverse_state_estimation}. The precise training details are described in the suppl. materials.

\textbf{Metrics}. To evaluate the quality of realized trajectories, we compute the average (over timesteps) displacement error (ADE) of a realized trajectory compared to the expert one. A key aspect throughout the experiments is route conditioning. In some experiments we purposefully do not condition the agent on any form of navigation and the action distributions from the policy are expected to be wide, to cover multiple reasonable trajectories. In those cases we realize multiple trajectories and report the minimum ADE among them (min ADE). This measures whether the learned policy can cover any reasonable expert trajectory \cite{montali2024waymo}. Whenever route conditioning is used and the agent knows the intended destination, we only realize a single trajectory and report its ADE. We also report the minimum overlap and offroad rates. They equal the proportion of scenarios in which at least one collision or offroad event occurs within the trajectory of lowest ADE. 

\subsection{Evaluating the Analytic Predictors}
\label{subsec: awms_eval}

\textbf{Optimal control.} For optimal control, we evaluate the policy head of our driving agent. Table \ref{table: reactive_apg_performance} shows that our trajectories obtained from rolling out a policy trained with APG are accurate. Compared to the previous APG baseline \cite{nachkov2024autonomous}, our training procedure improves ADE by 9\%. Here the agent knows the heading towards the final destination and all methods are directly comparable. The standard deviation of the ADE across 5 random seeds is $3.7 \times 10^{-4}$, hence variability plays almost no role.

\begin{table}[t]
    \small
    \centering
    \begin{tabular}[width=1\columnwidth]{ p{0.24\columnwidth} | p{0.11\columnwidth} p{0.14\columnwidth} p{0.16\columnwidth} } \toprule[1.5pt]
         \textbf{Model} & ADE $\downarrow$ & overlap $\downarrow$ & offroad $\downarrow$ \\
         \midrule[1pt]
         DQN & 9.8300 & 0.0650 & 0.0370 \\
         BC & 3.6000 & 0.1120 & 0.1360 \\
         Wayformer & 2.3800 & 0.1070 & 0.0790 \\
         APG (previous) & 2.0083 &  0.0800 & 0.0282 \\

         APG (ours) & \textbf{1.8121} & \textbf{0.0669} & \textbf{0.0263} \\

         \bottomrule[1.5pt]
    \end{tabular}
    \captionsetup{aboveskip=0.15cm, belowskip=-0.4cm}
    \caption{\textbf{APG performance}. The agent is conditioned on the heading towards the final destination. \emph{Takeaway}: our APG implementation outperforms RL, behavior cloning, sequence prediction methods, as well as the previous APG baseline.}
    \label{table: reactive_apg_performance}
\end{table}

Our second optimal control experiment validates not whether the agent can reach a specific destination, but whether it can drive similar to a human without being provided with an intended destination (i.e. no route conditioning). Here we realize multiple trajectories from the stochastic policy and report the best one. To greatly improve performance we have made an important architectural change compared to the baseline APG, which we now describe.

The baseline APG suffers from a Gaussian collapse because the policy, parametrized as a \emph{Gaussian mixture with 6 components}, samples a particular action \emph{from across all of them}. This causes the individual Gaussians to eventually stack on top of each other. Inspired by \cite{nayakanti2023wayformer} instead, we sample the action from only that Gaussian, whose mean will bring the agent closest to the next expert state. During training the gradients in the backward pass only reach this component (winner-take-all), which completely prevents the Gaussian mixture from losing its multimodality. Table \ref{table:gaussian_mixture} shows that as the number of trajectories rolled out increases, the best one improves in performance. This is evidence that the policy covers the expert trajectories well.

\begin{table}[t]
  \small
  \centering
    \begin{tabular}[width=1\columnwidth]{ p{0.12\columnwidth} | p{0.16\columnwidth} p{0.19\columnwidth} p{0.21\columnwidth} } \toprule[1.5pt]
         Rollouts & \makecell[tl]{min\\ADE $\downarrow$} & \makecell[tl]{min\\overlap $\downarrow$} & \makecell[tl]{min\\offroad $\downarrow$} \\
         \midrule[1pt]
         1 & 3.5725 & 0.2229 & 0.1224 \\
         4 & 2.0225 & 0.1350 & 0.1130 \\
         16 & 1.3361 & 0.0956 & 0.1056 \\
         32 & \textbf{1.1414} &  \textbf{0.0840} & \textbf{0.1030} \\
         \bottomrule[1.5pt]
    \end{tabular}
  \captionsetup{aboveskip=0.15cm, belowskip=-0.2cm}
  \caption{\textbf{Retaining the multimodality of the policy}. \emph{Takeaway}: with no route conditioning, the action distribution is wide. The more trajectories we sample, the closer is one of them to the historic one, which means the agent successfully learns action distributions similar to the human expert's.}
  \label{table:gaussian_mixture}
\end{table}

\begin{table}[t]
    \small
    \centering    
    \begin{tabular}{c@{\hskip 5pt} c@{\hskip 5pt} c@{\hskip 5pt} c@{\hskip 5pt} c@{\hskip 5pt} c}
        \toprule[1.5pt]
        \makecell{APG\\(ours)} & \makecell{Traffic\\BotsV1.5} & 
        \makecell{MVTE\\} &
        \makecell{CogniBOT \\ v1.5} & GUMP & 
        \makecell{Behavior\\GPT}  \\
        \midrule[1pt]
        \textbf{1.141} & 1.883 & 1.677 & 1.883 & 1.604 & 1.415 \\
        \bottomrule[1.5pt]
    \end{tabular}
    \captionsetup{aboveskip=0.15cm, belowskip=-0.45cm}
    \caption{\textbf{MinADE comparison with multi-agent methods}. All methods report the best of 32 modes. Ours is evaluated only on the ego-agent. \emph{Takeaway}: our APG implementation is competitive to SOTA methods, though differences in the experimental setup limit perfect comparison.}
    \label{table: comparison to other methods}
\end{table}

In this setting with no route conditioning, we can compare also to state-of-the-art methods from the Waymo Sim Agents challenge \cite{montali2024waymo}, in Table \ref{table: comparison to other methods}. They predict on all traffic participants -- not only the ego-vehicle -- which is different from our single-agent setup. Nonetheless, we find the comparison useful and our method very competitive. TrafficBots \cite{zhang2024trafficbots}, MVTE \cite{wang2023multiverse}, CogniBOT, GUMP, and BehaviorGPT \cite{zhou2024behaviorgpt} constitute strong transformer models focused on architectural innovations. Compared to them, our method has a simple recurrent architecture but is trained using the differentiable dynamics. We leave the adaptation of AWMs to multi-agent settings as future work.

\begin{figure*}[t]
    \centering
    \includegraphics[width=\textwidth]{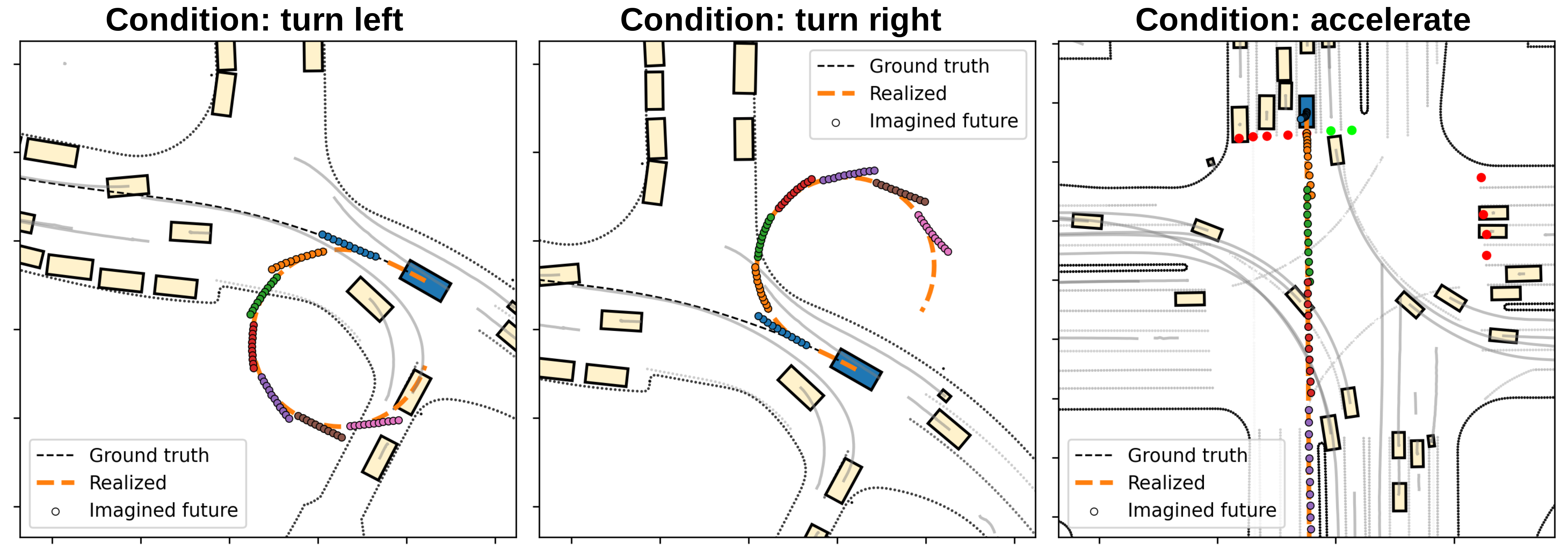}
    \captionsetup{belowskip=-0.35cm, aboveskip=0.1cm}
    \caption{\textbf{Predictions from the relative odometry.} We condition the ego-agent (\textcolor{RoyalBlue}{blue}) to go offroad, turn, or accelerate. The imagined trajectories, shown as scattered colored circles, represent the imagined future locations of the ego-vehicle in the next 1 second, plotted in different colors at the times 1s, 2s, ..., 7s throughout the episode. They align with the actual realized trajectory, which implies that the agent can imagine its future motion accurately. The ground truth historic trajectory is added for reference.}
    \label{fig: relative_odometry_trajectories}
\end{figure*}

\begin{figure*}[h]
    \centering
    \includegraphics[width=\textwidth]{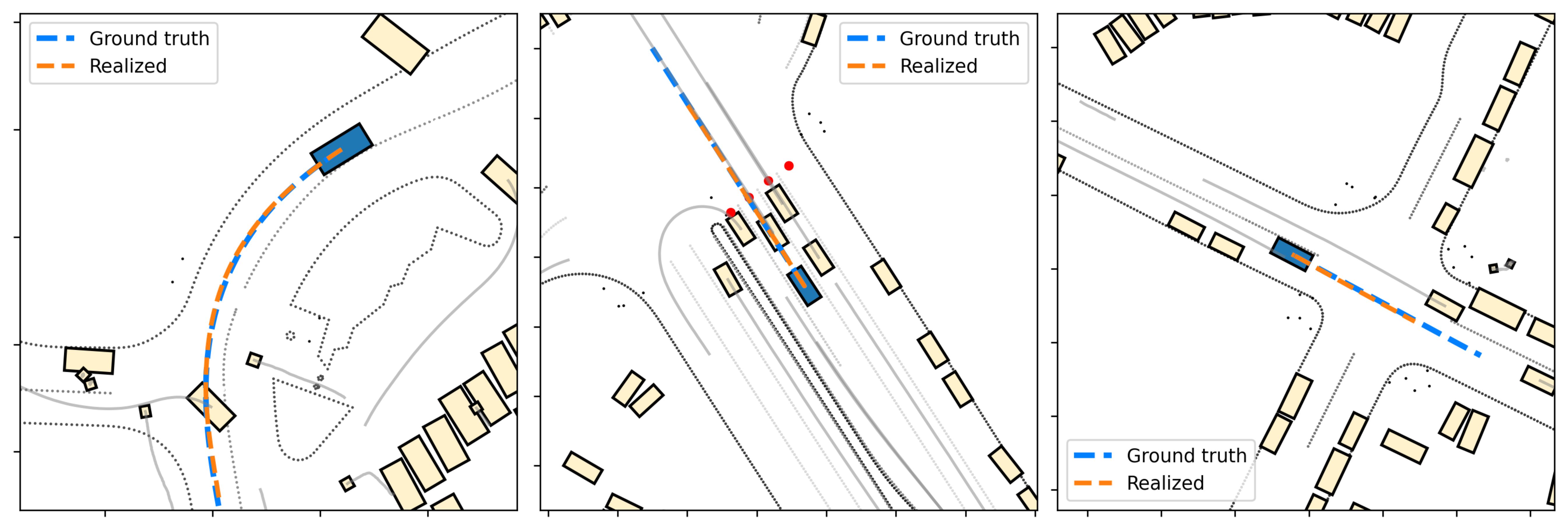}
    \captionsetup{belowskip=-0.35cm, aboveskip=0.2cm}
    \caption{\textbf{Trajectories obtained by using the optimal planner.} They are realistic and resemble those from the policy. Training such planners is possible due to the analytically available dynamics and inverse kinematics, to drive the action selection.}
    \label{fig: planner_trajectories}
\end{figure*}

\textbf{Relative odometry.} The main use of the agent's odometry head is to imagine the locations obtained from a sequence of ego-actions. Thus, to evaluate the odometry we measure how similar is the imagined future to the real one. We first produce qualitative results that demonstrate the odometric world model's controllability. To do so, we condition the agent's behavior by fixing its actions (instead of sampling them from the policy) to intentionally commit to a turn over a long time frame. Concurrently, the \emph{odometry predictor is used to autoregressively imagine} the next second (10 timesteps) of the planned motion, conditional on these fixed actions. We judge the imagined trajectory to be accurate if the imagination precisely aligns with the realized trajectory. Fig. \ref{fig: relative_odometry_trajectories} shows three examples. We observe accurate controllability -- if we condition the agent to turn left/right, accelerate/decelerate, the executed trajectory follows this motion, but importantly, the imagined trajectories for these action commands also represent similar motion. This means that the agent can accurately imagine the motion resulting from an action sequence.

\begin{table}[h]
    \small
    \centering
    \begin{tabular}[width=1\textwidth]{ p{0.25\columnwidth} | p{0.30\columnwidth} p{0.30\columnwidth} } \toprule[1.5pt]
         \textbf{Future steps} & With DiffSim & Without DiffSim \\
         \midrule[1pt]
         5 (0.5 sec) & 0.1698 & 0.3100 \\
         10 (1 sec) & 0.3475 & 0.7900 \\   
         15 (1.5 sec) & 0.5496 & 1.6200 \\       
         \bottomrule[1.5pt]
    \end{tabular}
    \captionsetup{aboveskip=0.15cm, belowskip=-0.4cm}
    \caption{\textbf{ADE [m] vs time horizon.} We measure the in-distribution (actions predicted by a trained policy) average distance between the imagined trajectory and the realized trajectory. \emph{Takeaway}: the accuracy of the imagination scales better with DiffSim, compared to without.}
    \label{table: odometry_accuracy}
\end{table}

Manually conditioning the predicted odometry on a desired action sequence could easily lead to out-of-distribution state-action sequences, as in Fig \ref{fig: relative_odometry_trajectories}. For example, driving offroad, making sudden sharp U-turns, or maximally accelerating can be considered rare events within the expert distribution. The accurate alignment between the imagined trajectory and the executed one shows that the network learns to generalize effectively. Nonetheless, scene complexity and large prediction lengths do limit the accuracy of the imagined trajectories. For in-distribution sequences where the actions are not set exogenously, but come from the policy, and for shorter future horizons, the odometry is very accurate, as shown in Table \ref{table: odometry_accuracy}.

\begin{table}[h]
    \small
    \centering
    \begin{tabular}[width=1\textwidth]{ p{0.28\columnwidth} | p{0.14\columnwidth} p{0.14\columnwidth} p{0.16\columnwidth} } \toprule[1.5pt]
         \textbf{Setting} & ADE $\downarrow$ & overlap $\downarrow$ & offroad $\downarrow$ \\
         \midrule[1pt]
         APG (previous) & 2.0083 &  0.0800 & 0.0282 \\
         Planner, Sec. \ref{subsection: optimal_planners} & \textbf{1.8734} &  \textbf{0.0719} & \textbf{0.0254} \\   
         \bottomrule[1.5pt]
    \end{tabular}
    \captionsetup{aboveskip=0.15cm, belowskip=-0.4cm}
    \caption{\textbf{Reactive evaluation of the planner.} Our design outperforms the previous APG method. \emph{Takeaway}: our planner, enabled by the corresponding AWM design, outperforms the previous APG baseline.}
    \label{table: planner_reactive}
\end{table}

\textbf{Optimal planners.} The optimal planner has a prescriptive role in that it predicts desired next states and the agent relies on the simulator's inverse kinematics to find the action that reaches them. We evaluate the planner with a combination of quantitative and qualitative metrics. Table \ref{table: planner_reactive} shows that obtaining an optimal planner using a differentiable environment is possible and even results in strong performance, improving over the baseline APG method on all metrics. For simplicity, the planner is deterministic and actions are chosen without any sampling. During the early stages of training, when the planner is still not very accurate, the resulting actions, obtained using the inverse kinematics, have large magnitude. The default inverse kinematics in Waymax clip them, which could prevent gradients from flowing back and could hinder training. Thus, we disable the action clipping in the inverse kinematics when training. At test time, we evaluate using the default inverse kinematics where action clipping is enabled.

Fig. \ref{fig: planner_trajectories} shows qualitative trajectories. They are smooth and realistic. When we condition the agent on the heading angle to the target, but not the distance to it, errors are mostly longitudinal and occur from over- and under-accelerating. Some turns are sharper but still follow reasonable trajectories.

\begin{figure}[t]
    \centering
    \includegraphics[width=1\columnwidth]{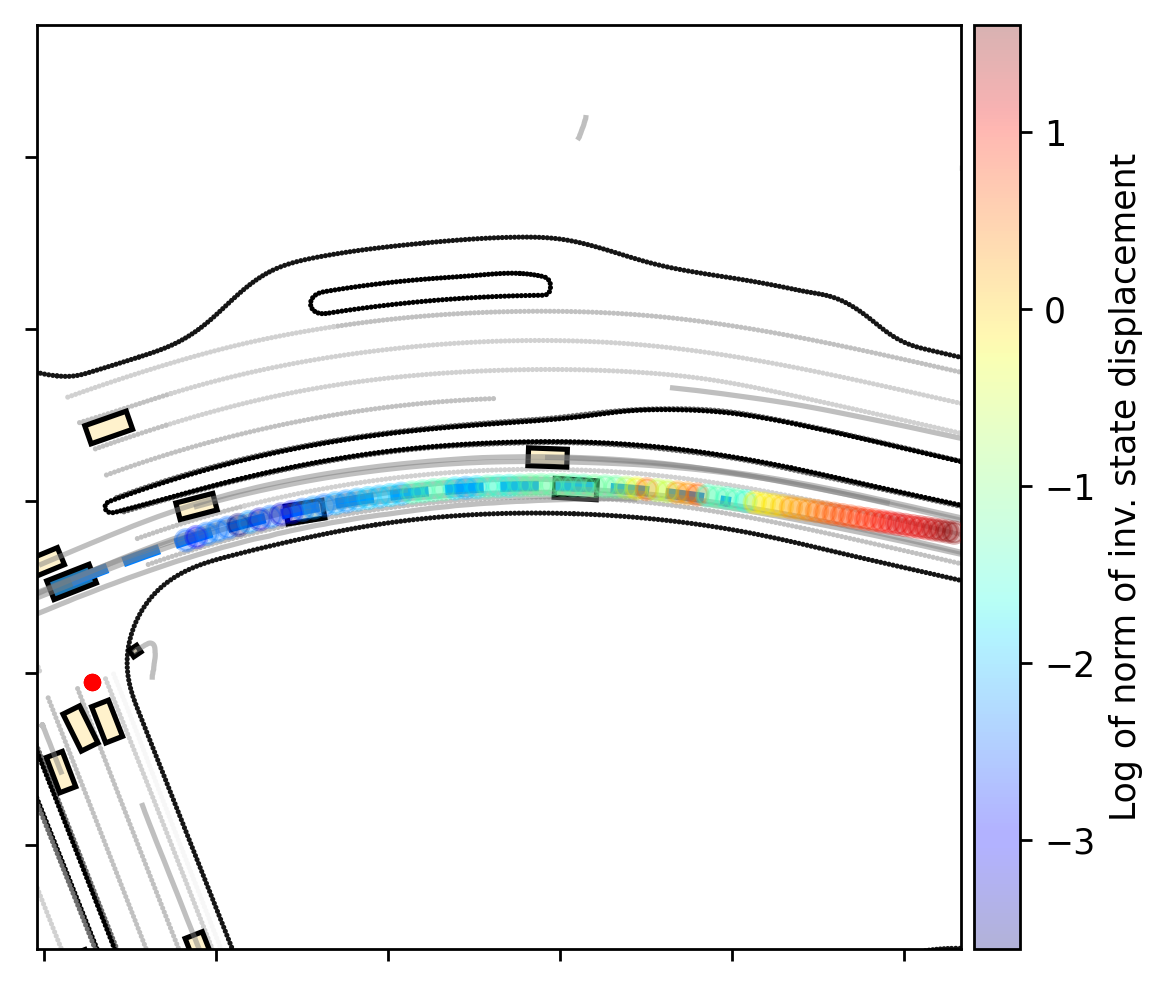}
    \captionsetup{belowskip=-0.5cm, aboveskip=0.00cm}
    \caption{\textbf{Realized trajectory colored according to the log-norm of the predicted inverse state displacements.} Since the ego-vehicle drives faster than the expert, the norm of the optimal inverse state predictions gradually increases.}
    \label{fig: inverse_state_preds}
\end{figure}

\textbf{Inverse optimal state prediction.} Finding a state in which a given action is optimal, is an inverse task. To motivate the setup for its evaluation, consider that if we start from an expert state $\hat{\mathbf{s}}_t$ and the selected action is optimal, $\mathbf{a}_t = \hat{\mathbf{a}}_t$, then the state we seek has a displacement of $\mathbf{0}$ from the given state $\hat{\mathbf{s}}_t$. Thus, if we assume the given actions are similar to the expert ones, the predicted displacement will indicate how far the ego-vehicle is from the current expert state $\hat{\mathbf{s}}_t$. The key quantity we look at is the norm of the predicted displacement $\lVert f_\phi^I(\mathbf{s}_t, \mathbf{a}_t) \rVert_2$. First we provide qualitative evaluation of this norm in Fig. \ref{fig: inverse_state_preds}. The results are meaningful -- as the agent over-accelerates, the optimal historical trajectory start lagging behind the realized one. The predicted inverse state displacement relative to the current state also increases. 

The norm of the predicted displacement can also be used as a confidence-based metric from which to select actions. In Table \ref{table: inverse_opt_state_as_rewards} we evaluate a model-predictive control setting (described next) where the agent selects those actions to execute that have the lowest predicted inverse state norm. Results show this setup is as accurate as when using the distance to the next log-state, which validates that the inverse state prediction can be used to select the right actions.

\begin{table}
    \small
    \centering
    \begin{tabular}[width=1\textwidth]{ p{0.32\columnwidth} | p{0.15\columnwidth} p{0.15\columnwidth} p{0.16\columnwidth} } \toprule[1.5pt]
         \textbf{Rewards}  & ADE $\downarrow$ & overlap $\downarrow$ & offroad $\downarrow$ \\
         \midrule[1pt]
         \makecell[l]{Negative distance\\to next log state}  & \textbf{1.8136} &  \textbf{0.0645} & 0.0226 \\
         \grayline
         \makecell[l]{Positive distance\\to next log state}  & 1.8247 &  0.0649 & 0.0229 \\
         \grayline
         \makecell[l]{Negative norm\\of inverse state} & 1.8138 &  0.0647 & \textbf{0.0218} \\
         \bottomrule[1.5pt]
    \end{tabular}
    \captionsetup{aboveskip=0.15cm, belowskip=-0.3cm}
    \caption{\textbf{Using the inverse state predictions.} \emph{Takeaway}: predictions from the inverse state estimation (last row) can be as useful as the explicit rewards for action selection.}
    \label{table: inverse_opt_state_as_rewards}
\end{table}

\subsection{Model Predictive Control (MPC)}
\label{subsec: mpc}

\begin{table}
    \small
    \centering
    \begin{tabular}[width=\textwidth]{ p{0.18\columnwidth} | p{0.12\columnwidth} | p{0.12\columnwidth} p{0.14\columnwidth} p{0.16\columnwidth} } \toprule[1.5pt]
        \makecell[tl]{\textbf{Rollouts},\\(top-$k$)} & \makecell[tl]{\textbf{Future}\\\textbf{steps}} & ADE~$\downarrow$ & overlap~$\downarrow$ &  offroad~$\downarrow$  \\
         \midrule[1.0pt]
             1 (1) & 1 &  3.5883 & 0.1842 & 0.0398 \\    
            4 (1) & 10 & 3.5975 & 0.1661 & 0.0348 \\ 
            8 (3) & 10 & 3.4719 & 0.1760 & 0.0369 \\  
            8 (3) & 20 & \textbf{3.2179} & 0.1576 & 0.0329 \\     
            12 (4) & 10 & 3.5258 & 0.1783 & 0.0369 \\ 
            12 (4) & 20 & 3.2486 & \textbf{0.1550} & \textbf{0.0318} \\  
         \bottomrule[1.5pt]
    \end{tabular}
    \captionsetup{aboveskip=0.15cm, belowskip=-0.4cm}
    \caption{\textbf{Model-predictive control at test time.} \emph{Takeaway}: increasing the imagined rollouts improves results.}
    \label{table: planning}
\end{table}

In the real world the agent can execute only one trajectory. Yet, the world models allow it to imagine multiple trajectories and select a single action refined from them. We use model-predictive control (MPC) as an experiment in this direction. At test time the agent uses the learned world modeling predictors to autoregressively imagine a number of future trajectories and to score them (according to their inverse state norms). The action to execute is obtained by aggregating the first actions from the $k$-best trajectories. Full details are provided in the suppl. materials. Importantly, this approach uses world modeling at test time to motivate the action selection. It goes beyond simple reactive decision-making because now the agent has to imagine the action's outcome and search for those actions with the best outcomes.

In Table \ref{table: planning} we condition on the expert heading so the agent knows the rough intended direction. Increasing the number of imagined trajectories (rollouts) and their lengths (future steps) improves performance compared to the reactive single-rollout case. Imagining 8 possible futures for the next 1 second (at 10 timesteps per second) improves over the reactive setting (1 rollout with 1 future timestep) by about 10\%. Thus, the world models allow the agent to play out multiple imagined trajectories and refine its actions from them. In this context, DiffSim enables us to have more varied and expressive world models, potentially leading to safer model-based action selection.
\section{Conclusion}
\label{sec: conclusion}

We have formulated three tasks that use differentiable simulation for world modeling. First, the relative odometry setup is \emph{predictive} -- it estimates the next state. Second, the optimal planner is \emph{prescriptive} -- it outputs a desired next state to visit. And third, the inverse optimal state allows for counterfactual estimation which can be interpreted as a useful confidence metric for the agent's own actions. We evaluated the corresponding Analytic World Models in diverse settings, including within a model-predictive control framework. Overall, we have shown that differentiable simulation unlocks the efficient learning of diverse world modeling predictors.

\section*{Acknowledgements}
This research was partially funded
by the Ministry of Education and Science of Bulgaria (support for INSAIT, part of the Bulgarian National Roadmap
for Research Infrastructure).

\bibliography{aaai2026}

@String(TOG= {ACM Trans. Graph.})

@String(TOG   = {ACM TOG})

@INPROCEEDINGS{nachkov2024autonomous,
  author={Nachkov, Asen and Paudel, Danda Pani and Van Gool, Luc},
  booktitle={2025 IEEE/RSJ International Conference on Intelligent Robots and Systems (IROS)}, 
  title={Autonomous Vehicle Controllers From End-to-End Differentiable Simulation}, 
  year={2025},
  volume={},
  number={}}

@article{gulino2024waymax,
  title={Waymax: An accelerated, data-driven simulator for large-scale autonomous driving research},
  author={Gulino, Cole and Fu, Justin and Luo, Wenjie and Tucker, George and Bronstein, Eli and Lu, Yiren and Harb, Jean and Pan, Xinlei and Wang, Yan and Chen, Xiangyu and others},
  journal={Advances in Neural Information Processing Systems},
  volume={36},
  year={2024}
}

@inproceedings{ettinger2021large,
  title={Large scale interactive motion forecasting for autonomous driving: The waymo open motion dataset},
  author={Ettinger, Scott and Cheng, Shuyang and Caine, Benjamin and Liu, Chenxi and Zhao, Hang and Pradhan, Sabeek and Chai, Yuning and Sapp, Ben and Qi, Charles R and Zhou, Yin and others},
  booktitle={Proceedings of the IEEE/CVF International Conference on Computer Vision},
  pages={9710--9719},
  year={2021}
}

@inproceedings{wiedemann2023training,
  title={Training efficient controllers via analytic policy gradient},
  author={Wiedemann, Nina and W{\"u}est, Valentin and Loquercio, Antonio and M{\"u}ller, Matthias and Floreano, Dario and Scaramuzza, Davide},
  booktitle={2023 IEEE International Conference on Robotics and Automation (ICRA)},
  pages={1349--1356},
  year={2023},
  organization={IEEE}
}

@article{freeman2021brax,
  title={Brax--a differentiable physics engine for large scale rigid body simulation},
  author={Freeman, C Daniel and Frey, Erik and Raichuk, Anton and Girgin, Sertan and Mordatch, Igor and Bachem, Olivier},
  journal={arXiv preprint arXiv:2106.13281},
  year={2021}
}

@article{montali2024waymo,
  title={The waymo open sim agents challenge},
  author={Montali, Nico and Lambert, John and Mougin, Paul and Kuefler, Alex and Rhinehart, Nicholas and Li, Michelle and Gulino, Cole and Emrich, Tristan and Yang, Zoey and Whiteson, Shimon and others},
  journal={Advances in Neural Information Processing Systems},
  volume={36},
  year={2024}
}

@article{wang2023multiverse,
  title={Multiverse Transformer: 1st Place Solution for Waymo Open Sim Agents Challenge 2023},
  author={Wang, Yu and Zhao, Tiebiao and Yi, Fan},
  journal={arXiv preprint arXiv:2306.11868},
  year={2023}
}

@article{sutton1999policy,
  title={Policy gradient methods for reinforcement learning with function approximation},
  author={Sutton, Richard S and McAllester, David and Singh, Satinder and Mansour, Yishay},
  journal={Advances in neural information processing systems},
  volume={12},
  year={1999}
}

@article{schrittwieser2020mastering,
  title={Mastering atari, go, chess and shogi by planning with a learned model},
  author={Schrittwieser, Julian and Antonoglou, Ioannis and Hubert, Thomas and Simonyan, Karen and Sifre, Laurent and Schmitt, Simon and Guez, Arthur and Lockhart, Edward and Hassabis, Demis and Graepel, Thore and others},
  journal={Nature},
  volume={588},
  number={7839},
  pages={604--609},
  year={2020},
  publisher={Nature Publishing Group UK London}
}

@inproceedings{codevilla2018end,
  title={End-to-end driving via conditional imitation learning},
  author={Codevilla, Felipe and M{\"u}ller, Matthias and L{\'o}pez, Antonio and Koltun, Vladlen and Dosovitskiy, Alexey},
  booktitle={2018 IEEE international conference on robotics and automation (ICRA)},
  pages={4693--4700},
  year={2018},
  organization={IEEE}
}

@article{hu2019difftaichi,
  title={Difftaichi: Differentiable programming for physical simulation},
  author={Hu, Yuanming and Anderson, Luke and Li, Tzu-Mao and Sun, Qi and Carr, Nathan and Ragan-Kelley, Jonathan and Durand, Fr{\'e}do},
  journal={arXiv preprint arXiv:1910.00935},
  year={2019}
}

@inproceedings{murthy2020gradsim,
  title={gradsim: Differentiable simulation for system identification and visuomotor control},
  author={Murthy, J Krishna and Macklin, Miles and Golemo, Florian and Voleti, Vikram and Petrini, Linda and Weiss, Martin and Considine, Breandan and Parent-L{\'e}vesque, J{\'e}r{\^o}me and Xie, Kevin and Erleben, Kenny and others},
  booktitle={International conference on learning representations},
  year={2020}
}

@inproceedings{nayakanti2023wayformer,
  title={Wayformer: Motion forecasting via simple \& efficient attention networks},
  author={Nayakanti, Nigamaa and Al-Rfou, Rami and Zhou, Aurick and Goel, Kratarth and Refaat, Khaled S and Sapp, Benjamin},
  booktitle={2023 IEEE International Conference on Robotics and Automation (ICRA)},
  pages={2980--2987},
  year={2023},
  organization={IEEE}
}

@article{laine2020modular,
  title={Modular primitives for high-performance differentiable rendering},
  author={Laine, Samuli and Hellsten, Janne and Karras, Tero and Seol, Yeongho and Lehtinen, Jaakko and Aila, Timo},
  journal={ACM Transactions on Graphics (ToG)},
  volume={39},
  number={6},
  pages={1--14},
  year={2020},
  publisher={ACM New York, NY, USA}
}

@article{sutton1991dyna,
  title={Dyna, an integrated architecture for learning, planning, and reacting},
  author={Sutton, Richard S},
  journal={ACM Sigart Bulletin},
  volume={2},
  number={4},
  pages={160--163},
  year={1991},
  publisher={ACM New York, NY, USA}
}

@inproceedings{pathak2017curiosity,
  title={Curiosity-driven exploration by self-supervised prediction},
  author={Pathak, Deepak and Agrawal, Pulkit and Efros, Alexei A and Darrell, Trevor},
  booktitle={International conference on machine learning},
  pages={2778--2787},
  year={2017},
  organization={PMLR}
}

@article{newbury2024review,
  title={A Review of Differentiable Simulators},
  author={Newbury, Rhys and Collins, Jack and He, Kerry and Pan, Jiahe and Posner, Ingmar and Howard, David and Cosgun, Akansel},
  journal={IEEE Access},
  year={2024},
  publisher={IEEE}
}

@article{zhang2019vr,
  title={Vr-goggles for robots: Real-to-sim domain adaptation for visual control},
  author={Zhang, Jingwei and Tai, Lei and Yun, Peng and Xiong, Yufeng and Liu, Ming and Boedecker, Joschka and Burgard, Wolfram},
  journal={IEEE Robotics and Automation Letters},
  volume={4},
  number={2},
  pages={1148--1155},
  year={2019},
  publisher={IEEE}
}

@inproceedings{dosovitskiy2017carla,
  title={CARLA: An open urban driving simulator},
  author={Dosovitskiy, Alexey and Ros, German and Codevilla, Felipe and Lopez, Antonio and Koltun, Vladlen},
  booktitle={Conference on robot learning},
  pages={1--16},
  year={2017},
  organization={PMLR}
}

@article{li2023dexdeform,
  title={Dexdeform: Dexterous deformable object manipulation with human demonstrations and differentiable physics},
  author={Li, Sizhe and Huang, Zhiao and Chen, Tao and Du, Tao and Su, Hao and Tenenbaum, Joshua B and Gan, Chuang},
  journal={arXiv preprint arXiv:2304.03223},
  year={2023}
}

@inproceedings{lutter2021differentiable,
  title={Differentiable physics models for real-world offline model-based reinforcement learning},
  author={Lutter, Michael and Silberbauer, Johannes and Watson, Joe and Peters, Jan},
  booktitle={2021 IEEE International Conference on Robotics and Automation (ICRA)},
  pages={4163--4170},
  year={2021},
  organization={IEEE}
}

@article{xu2021end,
  title={An end-to-end differentiable framework for contact-aware robot design},
  author={Xu, Jie and Chen, Tao and Zlokapa, Lara and Foshey, Michael and Matusik, Wojciech and Sueda, Shinjiro and Agrawal, Pulkit},
  journal={arXiv preprint arXiv:2107.07501},
  year={2021}
}

@article{lavington2024torchdriveenv,
  title={TorchDriveEnv: A Reinforcement Learning Benchmark for Autonomous Driving with Reactive, Realistic, and Diverse Non-Playable Characters},
  author={Lavington, Jonathan Wilder and Zhang, Ke and Lioutas, Vasileios and Niedoba, Matthew and Liu, Yunpeng and Green, Dylan and Naderiparizi, Saeid and Liang, Xiaoxuan and Dabiri, Setareh and {\'S}cibior, Adam and others},
  journal={arXiv preprint arXiv:2405.04491},
  year={2024}
}

@inproceedings{sun2022intersim,
  title={Intersim: Interactive traffic simulation via explicit relation modeling. In 2022 IEEE},
  author={Sun, Qiao and Huang, Xin and Williams, Brian C and Zhao, Hang},
  booktitle={RSJ International Conference on Intelligent Robots and Systems (IROS)},
  pages={11416--11423},
  year={2022}
}

@article{li2022metadrive,
  title={Metadrive: Composing diverse driving scenarios for generalizable reinforcement learning},
  author={Li, Quanyi and Peng, Zhenghao and Feng, Lan and Zhang, Qihang and Xue, Zhenghai and Zhou, Bolei},
  journal={IEEE transactions on pattern analysis and machine intelligence},
  volume={45},
  number={3},
  pages={3461--3475},
  year={2022},
  publisher={IEEE}
}

@article{martinez2017beyond,
  title={Beyond grand theft auto V for training, testing and enhancing deep learning in self driving cars},
  author={Martinez, Mark and Sitawarin, Chawin and Finch, Kevin and Meincke, Lennart and Yablonski, Alex and Kornhauser, Alain},
  journal={arXiv preprint arXiv:1712.01397},
  year={2017}
}

@article{vinitsky2022nocturne,
  title={Nocturne: a scalable driving benchmark for bringing multi-agent learning one step closer to the real world},
  author={Vinitsky, Eugene and Lichtl{\'e}, Nathan and Yang, Xiaomeng and Amos, Brandon and Foerster, Jakob},
  journal={Advances in Neural Information Processing Systems},
  volume={35},
  pages={3962--3974},
  year={2022}
}

@book{bertsekas2012dynamic,
  title={Dynamic programming and optimal control: Volume I},
  author={Bertsekas, Dimitri},
  volume={4},
  year={2012},
  publisher={Athena scientific}
}

@article{arroyo2022reinforced,
  title={Reinforced model predictive control (RL-MPC) for building energy management},
  author={Arroyo, Javier and Manna, Carlo and Spiessens, Fred and Helsen, Lieve},
  journal={Applied Energy},
  volume={309},
  pages={118346},
  year={2022},
  publisher={Elsevier}
}

@inproceedings{romero2024actor,
  title={Actor-critic model predictive control},
  author={Romero, Angel and Song, Yunlong and Scaramuzza, Davide},
  booktitle={2024 IEEE International Conference on Robotics and Automation (ICRA)},
  pages={14777--14784},
  year={2024},
  organization={IEEE}
}

@article{toussaint2018differentiable,
  title={Differentiable physics and stable modes for tool-use and manipulation planning},
  author={Toussaint, Marc A and Allen, Kelsey Rebecca and Smith, Kevin A and Tenenbaum, Joshua B},
  year={2018},
  publisher={Robotics: Science and systems foundation}
}

@article{holl2020learning,
  title={Learning to control pdes with differentiable physics},
  author={Holl, Philipp and Koltun, Vladlen and Thuerey, Nils},
  journal={arXiv preprint arXiv:2001.07457},
  year={2020}
}

@article{ha2018world,
  title={World models},
  author={Ha, David and Schmidhuber, J{\"u}rgen},
  journal={arXiv preprint arXiv:1803.10122},
  year={2018}
}

@article{moerland2023model,
  title={Model-based reinforcement learning: A survey},
  author={Moerland, Thomas M and Broekens, Joost and Plaat, Aske and Jonker, Catholijn M and others},
  journal={Foundations and Trends{\textregistered} in Machine Learning},
  volume={16},
  number={1},
  pages={1--118},
  year={2023},
  publisher={Now Publishers, Inc.}
}

@article{degrave2019differentiable,
  title={A differentiable physics engine for deep learning in robotics},
  author={Degrave, Jonas and Hermans, Michiel and Dambre, Joni and Wyffels, Francis},
  journal={Frontiers in neurorobotics},
  volume={13},
  pages={6},
  year={2019},
  publisher={Frontiers Media SA}
}

@article{heiden2021disect,
  title={Disect: A differentiable simulation engine for autonomous robotic cutting},
  author={Heiden, Eric and Macklin, Miles and Narang, Yashraj and Fox, Dieter and Garg, Animesh and Ramos, Fabio},
  journal={arXiv preprint arXiv:2105.12244},
  year={2021}
}

@article{xu2022accelerated,
  title={Accelerated policy learning with parallel differentiable simulation},
  author={Xu, Jie and Makoviychuk, Viktor and Narang, Yashraj and Ramos, Fabio and Matusik, Wojciech and Garg, Animesh and Macklin, Miles},
  journal={arXiv preprint arXiv:2204.07137},
  year={2022}
}

@article{song2024learning,
  title={Learning Quadruped Locomotion Using Differentiable Simulation},
  author={Song, Yunlong and Kim, Sangbae and Scaramuzza, Davide},
  journal={arXiv preprint arXiv:2403.14864},
  year={2024}
}

@article{geilinger2020add,
  title={Add: Analytically differentiable dynamics for multi-body systems with frictional contact},
  author={Geilinger, Moritz and Hahn, David and Zehnder, Jonas and B{\"a}cher, Moritz and Thomaszewski, Bernhard and Coros, Stelian},
  journal={ACM Transactions on Graphics (TOG)},
  volume={39},
  number={6},
  pages={1--15},
  year={2020},
  publisher={ACM New York, NY, USA}
}

@misc{warp2022,
title= {Warp: A High-performance Python Framework for GPU Simulation and Graphics},
author = {Miles Macklin},
month = {March},
year = {2022},
note= {NVIDIA GPU Technology Conference (GTC)},
howpublished = {\url{https://github.com/nvidia/warp}}
}

@article{de2018end,
  title={End-to-end differentiable physics for learning and control},
  author={de Avila Belbute-Peres, Filipe and Smith, Kevin and Allen, Kelsey and Tenenbaum, Josh and Kolter, J Zico},
  journal={Advances in neural information processing systems},
  volume={31},
  year={2018}
}

@article{lin2022diffskill,
  title={Diffskill: Skill abstraction from differentiable physics for deformable object manipulations with tools},
  author={Lin, Xingyu and Huang, Zhiao and Li, Yunzhu and Tenenbaum, Joshua B and Held, David and Gan, Chuang},
  journal={arXiv preprint arXiv:2203.17275},
  year={2022}
}

@article{heeg2024learning,
  title={Learning Quadrotor Control From Visual Features Using Differentiable Simulation},
  author={Heeg, Johannes and Song, Yunlong and Scaramuzza, Davide},
  journal={arXiv preprint arXiv:2410.15979},
  year={2024}
}

@article{zhang2024trafficbots,
  title={Trafficbots v1. 5: Traffic simulation via conditional vaes and transformers with relative pose encoding},
  author={Zhang, Zhejun and Sakaridis, Christos and Van Gool, Luc},
  journal={arXiv preprint arXiv:2406.10898},
  year={2024}
}

@article{zhou2024behaviorgpt,
  title={Behaviorgpt: Smart agent simulation for autonomous driving with next-patch prediction},
  author={Zhou, Zikang and Haibo, HU and Chen, Xinhong and Wang, Jianping and Guan, Nan and Wu, Kui and Li, Yung-Hui and Huang, Yu-Kai and Xue, Chun Jason},
  journal={Advances in Neural Information Processing Systems},
  volume={37},
  pages={79597--79617},
  year={2024}
}

@article{li2023scenarionet,
  title={Scenarionet: Open-source platform for large-scale traffic scenario simulation and modeling},
  author={Li, Quanyi and Peng, Zhenghao Mark and Feng, Lan and Liu, Zhizheng and Duan, Chenda and Mo, Wenjie and Zhou, Bolei},
  journal={Advances in neural information processing systems},
  volume={36},
  pages={3894--3920},
  year={2023}
}

\end{document}